\documentclass[letterpaper]{article} 
\usepackage{aaai24}  
\usepackage{times}  
\usepackage{helvet}  
\usepackage{courier}  
\usepackage[hyphens]{url}  
\usepackage{graphicx} 
\usepackage{natbib}  
\usepackage{caption} 
\usepackage{algorithm}
\usepackage{algorithmic}
\usepackage{amsfonts}
\usepackage{amsmath}

\usepackage{newfloat}
\usepackage{listings}
\DeclareCaptionStyle{ruled}{labelfont=normalfont,labelsep=colon,strut=off} 
\floatstyle{ruled}
\newfloat{listing}{tb}{lst}{}
\floatname{listing}{Listing}
\usepackage[USenglish]{babel}
\usepackage{microtype}
\usepackage{booktabs}
\usepackage[noabbrev,capitalize]{cleveref}
\usepackage{csquotes}
\usepackage{todonotes}
\usepackage{multirow}

\newcommand{\ThreeDifFusionDet}{3DifFusionDet}
\newcommand{\LiDAR}{LiDAR}

\title{{\ThreeDifFusionDet}: Diffusion Model for 3D Object Detection with Robust {\LiDAR}-Camera Fusion}
\author{
    Xinhao Xiang\textsuperscript{\rm 1}, Simon Dr{\"a}ger\textsuperscript{\rm 1}, Jiawei Zhang\textsuperscript{\rm 1}  
}

\affiliations{
    \textsuperscript{\rm 1}IFM Lab, University of California, Davis \\
    \texttt{xinhao@ifmlab.org}, 
    \texttt{sdraeger@ucdavis.edu}, 
    \texttt{jiawei@ifmlab.org}
}

\begin{document}

\maketitle

\begin{abstract}


Good 3D object detection performance from {\LiDAR}-Camera sensors demands seamless feature alignment and fusion strategies.  
We propose the {\ThreeDifFusionDet} framework in this paper, which structures 3D object detection as a denoising diffusion process from noisy 3D boxes to target boxes. 
In this framework, ground truth boxes diffuse in a random distribution for training, and the model learns to reverse the noising process. 
During inference, the model gradually refines a set of boxes that were generated at random to the outcomes.
Under the feature align strategy, the progressive refinement method could make a significant contribution to robust {\LiDAR}-Camera fusion. 
The iterative refinement process could also demonstrate great adaptability by applying the framework to various detecting circumstances where varying levels of accuracy and speed are required. 
Extensive experiments on KITTI, a benchmark for real-world traffic object identification, revealed that {\ThreeDifFusionDet} is able to perform favorably in comparison to earlier, well-respected detectors.
\end{abstract}

\section{Introduction}

With the introduction of 3D sensors and a variety of 3D understanding applications, 3D recognition~\cite{ObjectNet3D}, object detection~\cite{graikos2023diffusion}, and segmentation~\cite{PointNet} have come under more scrutiny in research.
3D data is crucial for a wide range of applications, including navigation~\cite{Kitti, PL-SVO}, augmented reality~\cite{Multiple3DObjectTrackingforAugmentedReality}, and robotics~\cite{RichFeatureHierarchies}.
Among these tasks, 3D object detection is a critical issue and a key stage in many pipelines for 3D comprehension. 
Unlike traditional 2D object detection, 3D detection provides richer spatial information about objects, allowing for accurate depth perception and volumetric understanding.
Those advantages are crucial in applications such as autonomous driving, where discerning the precise distance and orientation of surrounding vehicles is an essential matter of safety.

Most high-performance 3D object detectors nowadays utilize several sensors, including cameras and {\LiDAR}, combining information to perform 3D object detection~\cite{MV3D, MVXNet, LIFT}.
Combining the high-resolution visual cues from cameras with the depth information from {\LiDAR} provides a more comprehensive scene representation, enhancing detection accuracy, especially in complex environments.
Moreover, multi-modal fusion mitigates each sensor's limitations~\cite{TransFuser,TransFusion}. It improves the detection system's robustness, making it less susceptible to errors or ambiguities from relying on a single sensor modality~\cite{3DODinAD}.
Owing to these significant advantages, multi-modal fusion is arguably the field's future trajectory.

There are two crucial elements worth investigating, the first being the 3D detecting head. 
Pre-defined anchor-box proposals~\cite{FrustumNet, SECOND, PartA2Point} and learnable anchor-free queries~\cite{PointRCNN,3DSSD, VoxelNeXt} are examples of traditional methods.
The first group of methods involves a pre-defined set of 3D bounding boxes, called anchors, of different shapes and sizes that slide across the spatial dimensions of the feature map.
For the second group, instead of using pre-defined anchor boxes, these methods predict objects directly from feature points or use other mechanisms, such as sparse convolution windows~\cite{VoxelNeXt, SWFormer} or point bases~\cite{3DSSD}.  

Originating from the 2D object detection field, the query-based detection paradigm has recently received a lot of attention~\cite{SMCA, DN-DETR, SparseRCNN, DeformableDETR}, thanks to DETR's~\cite{DETR} proposal of learnable object queries to do away with the hand-designed components and build up an end-to-end detection pipeline.
Several attempts in 3D object detection have been conducted~\cite{3DETR, TransFusion}. 
These works manage a straightforward and efficient architecture but still depend on a predetermined set of learnable equations.

More recently, after observing tremendous success in several generation tasks~\cite{StructuredDenoisingDiffusionModel, BlendedDiffusionforText_drivenEditing, DiffusionforMoleculeGeneration, DiffusionProbabilisticModelingOfProteinBackbones}, diffusion models have been investigated in perception tasks like image segmentation~\cite{Diffusioninst, EfficientSS, AGeneralistFramework,graikos2023diffusion}, text-video retrieval~\cite{DiffusionRet}, human pose estimation~\cite{Diffusion_Based3DHumanPoseEstimation}.
DiffusionDet~\cite{DiffusionDet} proposes a novel framework that directly detects objects from a set of random boxes by using a diffusion model~\cite{DDPM}.
The underlying principle of their noise-to-box paradigm is analogous to the noise-to-image procedure observed in the denoising diffusion models~\cite{DiffusionModelsBeatGANs, DDPM}.
These models construct images by iteratively eliminating noise through a trained denoising mechanism, achieving good but improvable performance in 2D detection.

In this paper, \textit{we extend the usage of noise-to-image denoising pipelines to 3D object detection.}
To exploit its potential benefits and performance as much as possible, we design our noise-to-3DBox paradigm under a {\LiDAR}-camera fusion framework, which can provide a richer, more robust, and comprehensive detection paradigm.
We structure 3D object detection as a denoising diffusion process~\cite{DDPM} from noisy 3D bounding boxes to target boxes.
In this framework, ground truth boxes diffuse in a Gaussian distribution for training. These noisy boxes extract Regions of Interest (RoI;~\cite{Mask_RCNN, Faster_RCNN, MV3D, PartA2Point}) from the output feature map of the backbone encoder.
These RoI features are forwarded to the detection decoder, trained to estimate the noise-free ground truth boxes by learning the reverse process.
When the number of diffusion steps \(D \in \mathbb{N}\) is large enough, e.g., \ \(D = 1000\), the noisy boxes can be viewed as random variables sampled from the space of bounding boxes.
During inference, several randomly generated bounding boxes are sampled for the learned reversing process to predict the 3D ground truth boxes.

The second element is the camera-{\LiDAR} fusion alignment strategies.
Camera images and LiDAR cloud features are two inherently different features, where images are a dense data representation with most of the space in a point cloud being empty (\enquote{sparse}).
Images also contain rich texture and color information while lacking depth information.
On the other hand, simply executing the RoIAlign operations~\cite{Mask_RCNN, DiffusionDet} on the fused features does not fully use the complementary information the two modalities provide.
Drawing upon this, besides putting the point cloud RoIAlign under the encoded 3D features, we also create a second branch that performs an image RoIAlign under the encoded 2D features.
As for the connection of these two feature branches, simply concatenating them will suffer from information cuts which lead to reduced performance.
To this end, a multi-head cross attention mechanism~\cite{Transformers} is introduced to align these features deeply.
These aligned features are sent to the detection head in order to predict the ground truth boxes without noise.

We evaluate {\ThreeDifFusionDet} on the KITTI 3D object detection benchmark~\cite{Kitti}.
With proper feature extraction and fusion backbones~\cite{ResNet, MVXNet, VoxelNet, VoxelNeXt}, 3DifFusionDet achieves higher mean Average Precision (mAP), which outperforms several latest and popular methods.
Different from all other SOTA 3D object detection methods, our {\ThreeDifFusionDet} has the ability to perform multi-step inference once training has concluded.
Besides that, our framework has the flexibility to reach different levels of detection accuracy and inference speed by changing the number of denoising sampling steps, both of which reveal broader potential usages.  

Our contributions can be summarized as follows:
\begin{itemize}
    \item We formulate 3D object detection as a generative denoising process and propose {\ThreeDifFusionDet}, which to the best of our knowledge is the first study to apply the diffusion model to 3D object detection.
    \item We investigate the best camera-{\LiDAR} fusion alignment strategies under the generative denoising process framework and propose 2-branch fusion align strategies to exploit the complementary information that the two modalities provide. 
    \item Extensive experiments are conducted on the KITTI benchmark.
    {\ThreeDifFusionDet} achieves competitive results compared with existing well-designed approaches, showing a promising future of diffusion models in 3D object detection tasks.
\end{itemize}

\begin{figure*}
    \includegraphics[width=\textwidth]{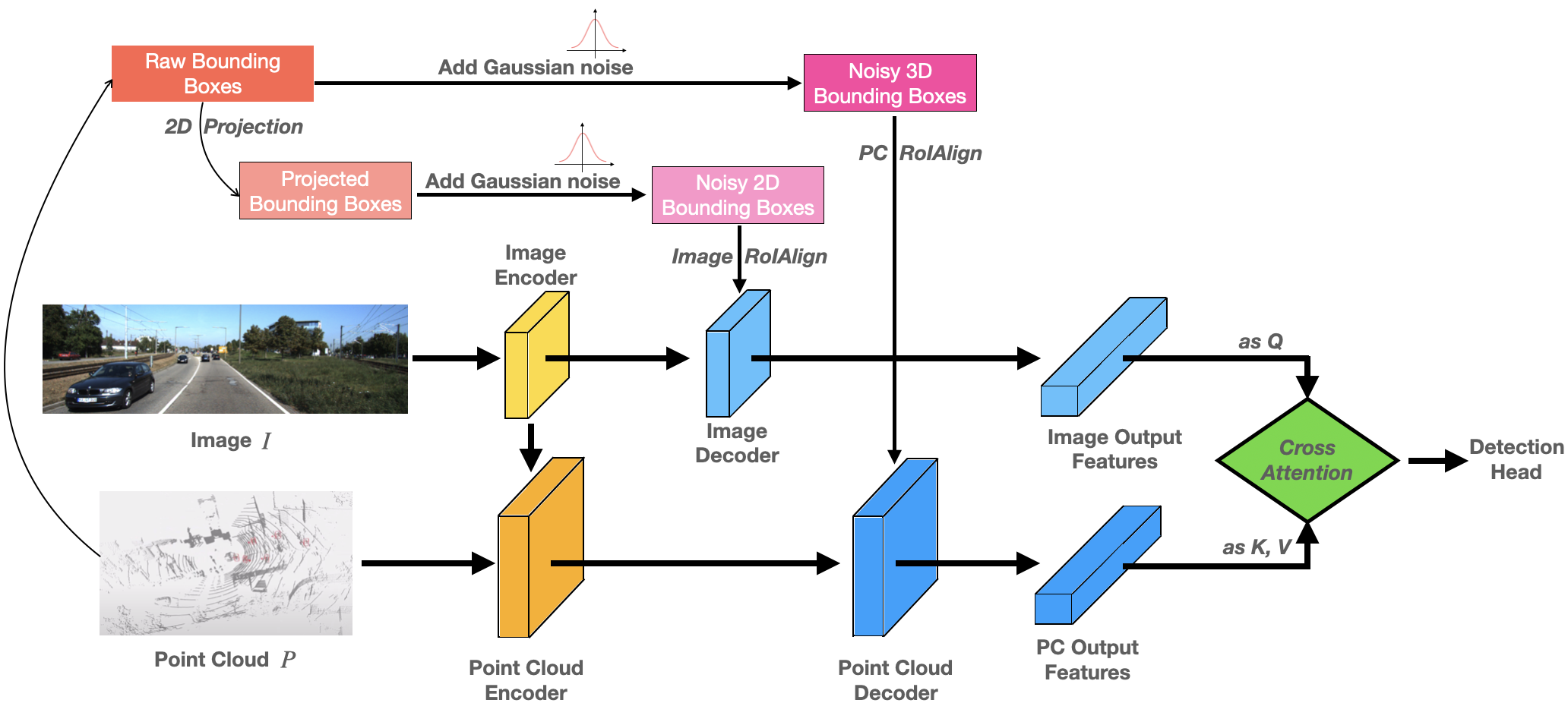}
    \caption{Overview of {\ThreeDifFusionDet}.}
    \label{fig:fig1}
\end{figure*}

\section{Related Work}

\subsection{3D Object Detection with {\LiDAR}-Camera Fusion}

For 3D object detection, camera and {\LiDAR} are two complementing sensor types.
{\LiDAR} sensors specialize in 3D localization and provide rich information about 3D structures, while cameras provide color information from which rich semantic features can be derived~\cite{3DODinAD}.
Many efforts have been made to accurately detect 3D objects by fusing data from cameras and {\LiDAR}s.
State-of-the-art approaches~\cite{MultiModalFusionTransformer, TransFusion, LIFT, BEVFusion, UniDistill} are primarily based on {\LiDAR}-based 3D object detectors and strive to incorporate image information into various stages of a {\LiDAR} detection pipeline since {\LiDAR}-based detection methods perform significantly better than camera-based methods. 
Combining the two modalities necessarily increases computing cost and inference time lag due to the complexity of {\LiDAR}-based and camera-based detection systems.
As a result, the problem of effectively fusing information from several modes still exists.

\subsection{Diffusion Models}

A diffusion model~\cite{DiffusionModel} is a generative model that gradually deconstructs observed data by introducing noise and restoring the original data by reversing this process.
Diffusion models and denoising score matching are connected via denoising diffusion probabilistic models~\cite{DDPM}, which have recently sparked interest in applications of computer vision~\cite{EfficientSS,Diffusioninst,DiffusionDet}.
In several fields such as graph generation~\cite{DiffusionProbabilisticModelingOfProteinBackbones,DiffusionforMoleculeGeneration}, language understanding~\cite{Diffusion_LMImprovesControllableTextGeneration,StructuredDenoisingDiffusionModelsinDiscreteState_Spaces}, robust learning~\cite{GuidedDiffusionModelforAdversarialPurification,DiffusionModelsforAdversarialPurification} and temporal data modeling~\cite{NeuralMarkovControlled_SDE,DiffWave}
For instance, DDPMs have been used for super-resolution applications by SR3~\cite{ImageSuperResolutionviaIterativeRefinement}.

\subsection{Diffusion Models in Vision Tasks}

Diffusion models have achieved great success in image generation and synthesis~\cite{Diffusioninst,DiffusionModelsBeatGANs,DDPM,GenerativeModelingByEstimatingGradientsOfTheDataDistribution}.
Some pioneer works adopt the diffusion model for image segmentation tasks~\cite{Diffusioninst,SegDiff,EfficientSS,DecoderDenoisingPretrainingForSemanticSegmentation}.
Compared to these fields, their potential for object detection has yet to be fully explored.
Previous approaches using a diffusion model for object detection~\cite{DiffusionDet} are restricted to 2D bounding boxes.
Compared to 2D detection, 3D detection provides richer spatial information about objects, allowing for accurate depth perception and volumetric understanding, making it crucial in applications such as autonomous driving, where discerning the precise distance and orientation of surrounding vehicles is an essential aspect of safety.
To the best of our knowledge, this is the first work that adopts a diffusion model to achieve 3D object detection.


\section{Proposed Method}


\paragraph{Notation and Terminology.}
In the remainder of this paper, we use upper or lower case letters (e.g., \(X\) or \(x\)) to represent scalars, lower case bold letters (e.g., \(\mathbf{x}\)) to denote column vectors, and bold-face upper case letters (e.g., \(\mathbf{X}\)) to represent matrices, and upper case calligraphic letters (e.g., \(\mathcal{X}\)) to indicate sets or higher-order tensors.
We use \(\mathbf{X}^\top\) and \(\mathbf{x}^\top\) to represent the transpose of any matrix \(\mathbf{X}\) and vector \(\mathbf{x}\).


\subsection{Preliminaries}

\paragraph{Diffusion Models.}
Modern diffusion models often employ two Markov chains: a forward chain to corrupt the image using noise and a reverse chain to refine the noise back into an image~\cite{DDPM}.
Formally, the forward noise perturbing process at time step \(t\) is defined as \(q\left(\mathbf{x}_t \mid \mathbf{x}_{t-1}\right)\) for a data distribution \(\mathbf{x}_0 \sim q\left(\mathbf{x}_0\right)\).
A variance schedule \(\beta_1, \cdots, \beta_T\) is used to progressively add, customarily Gaussian, noise to the data:

\begin{equation}
    q\left(\mathbf{x}_t \mid \mathbf{x}_{t-1}\right)=\mathcal{N}\left(\mathbf{x}_t ; \sqrt{1-\beta_t} \mathbf{x}_{t-1}, \beta_t \mathbf{I}\right).
\end{equation}

Given \(\mathbf{x}_0\), we can quickly generate a sample of \(\mathbf{x}_t\) by sampling a Gaussian random vector \(\boldsymbol{\epsilon} \sim \mathcal{N}(\mathbf{0}, \mathbf{I})\) and performing the following transformation:
\begin{equation}
    \mathbf{x}_t=\sqrt{\bar{\alpha}_t} \mathbf{x}_0+\left(1-\bar{\alpha}_t\right) \boldsymbol{\epsilon}
\end{equation}
where \(\bar{\alpha}_t=\prod_{s=0}^t\left(1-\beta_s\right)\).

When being trained, a neural network is taught to forecast \(\mathbf{x}_0\) from \(\mathbf{x}_t\) for various \(t \in \{1, \dots, T\}\).
When making inferences, we begin with random noise \(\mathbf{x}_T\) and iteratively employ the reverse chain to produce \(\mathbf{x}_0\).

\paragraph{DiffusionDet.}
It is the first diffusion model for the 2D object detection problem~\cite{DiffusionDet}.
In their context, data samples are a collection of bounding boxes according to the formula \(\mathbf{x}_0 = \mathbf{b}\), where \(\mathbf{b} \in \mathcal{R}^{N \times 4}\) is a set of \(N\) boxes.
DiffusionDet builds the diffusion process during training, then learns to reverse it, by training a neural network \(f_\theta\left(\mathbf{x_t}, t\right)\) to predict \(\boldsymbol{x}_0\) from \(\boldsymbol{x}_t\) by minimizing the training objective with \(\ell_2\) loss~\cite{l2Loss}:

\begin{equation}
    \mathcal{L}_{\text{train}}=\frac{1}{2}\left\|f_\theta\left(\mathbf{x}_t, t\right)-\mathbf{x}_0\right\|^2.
\end{equation}

The model can handle a fixed number of 2D instance boxes by padding additional boxes onto the initial ground truth boxes.
DiffusionDet is optimized using set prediction loss~\cite{DETR} with optimum transport assignment~\cite{OTA} serving as the label assignment approach.

For the inference approach, DiffusionDet additionally makes use of DDIM~\cite{DDIM} to improve the boxes for the subsequent iteration of the sampling process.

\subsection{{\ThreeDifFusionDet} Architecture}

\paragraph{Framework Overview.}
\Cref{fig:fig1} shows the overall architecture of the proposed {\ThreeDifFusionDet}.
It accepts multi-modal inputs, which include both RGB images and point clouds.
The input images are defined as \(\mathcal{I} \in \mathbb{R}^{H_I \times W_I \times 3}\), where \(H_I\) and \(W_I\) denote the image height and width dimensions, respectively.
Meanwhile, the input point cloud is represented as a set of \(3 \mathrm{D}\) points \(\mathcal{P} \in \mathbb{R}^{H_P \times W_P \times D_P}\) in the \(H_P \cdot W_P \cdot D_P\) 3D space, where each point is a vector of its \((x, y, z)\)-coordinate.

We divide the entire model into feature extraction and feature decoding components for the same computationally intractable reason as DiffusionDet~\cite{DiffusionDet}: it would be (even more) difficult to directly apply \(f_\theta\) on the raw 3D features at each iteration step.
The feature extraction part runs only once to extract a deep feature representation from the raw input \(\mathcal{X}\), while the feature decoding component takes this deep feature as a condition and trains to progressively draw the box predictions from noisy boxes \(\mathbf{u_t}\).

To make full use of the complementary information provided by the two modalities, we separate encoders and decoders for each modality.
In addition, we generate the noisy boxes \(\mathbf{u^{i}_t}\) and \(\mathbf{u^{p}_t}\) distinctly using the diffusion model, training the image decoder and point cloud decoder individually to refine the 2D and 3D features.
As for the connection of these two feature branches, simply concatenating them is going to incur an information cut which leads to reduced performance.
To this end, a multi-head cross-attention mechanism~\cite{Attention} is introduced to deeply align these features.
These aligned features are sent to the detection head to predict the final ground truth boxes without noise.
We introduce those afore-mentioned functional components in detail to the reader:

\paragraph{Feature Extraction and Fusion Modules.}
Given raw images \(\mathbf{I} \in \mathbb{R}^{H_I \times W_I \times 3}\) and 3D points \(\mathcal{P} \in \mathbb{R}^{H_P \times W_P \times D_P}\), we use a separate feature extractor to encode them.
For the image encoder, following \citet{DiffusionDet}, we try convolutional neural networks like ResNet~\cite{ResNet} in the implementation of {\ThreeDifFusionDet}.
In light of \citet{FPN}, ResNet generated using a Feature Pyramid Network. 

For the point encoder, we utilize voxel-based methods~\cite{MVXNet, VoxelNeXt} to extract, and adopt sparse-based methods~\cite{SECOND, FSD++, SST} to process.
Voxel-based methods convert {\LiDAR} points to voxels.
Compared to other families of point feature extraction methods, such as point-based methods, these methods discretize the point cloud into equally spaced 3D grids, reducing memory needs while keeping the original 3D shape information as much as possible.
Sparsity-based processing methods further help networks to be computationally more efficient.
These benefits balance out the diffusion model's comparably high computational requirements. 

Compared to 2D features, 3D features contain an additional dimension, making learning more challenging.
Considering this, besides feature extraction from raw modalities, we add a fusing path that adds the extracted image features as another input for the point encoder, promoting information exchange as well as exploiting learning from more diverse sources.
Here we adopt the PointFusion strategy from \citet{MVXNet}, where points from the {\LiDAR} sensor are projected onto the image plane.
The concatenation of image features and the corresponding points are then jointly processed by the VoxelNet architecture.

\paragraph{Feature Decoders.}
The extracted image features \(\mathbf{f^{i}}\) and the extracted point features \(\mathbf{f^{p}}\) serve as input for the corresponding image and point decoders.
Each decoder additionally incorporates input from the distinctively created noisy boxes, \(\mathbf{u^{i}_t}\) or \(\mathbf{u^{p}_t}\), to learn to refine the 2D and 3D features separately in addition to the corresponding extracted features.
How these noisy boxes are created will be introduced in detail in \cref{sec:training-inference}.

The image decoder, taking inspiration from Sparse~R-CNN~\cite{SparseRCNN}, receives input from a collection of 2D proposal boxes to crop RoI-features~\cite{Fast_RCNN} from feature maps created by the image encoder.
The point decoder receives input from a collection of 3D proposal boxes to crop RoI features~\cite{PartA2Point, MV3D, AVOD} from feature maps created by the image encoder.
For the point decoder, the input is a set of 3D proposal boxes to crop 3D RoI-feature from feature maps generated by the point encoder.

\begin{figure*}
    \includegraphics[width=\textwidth]{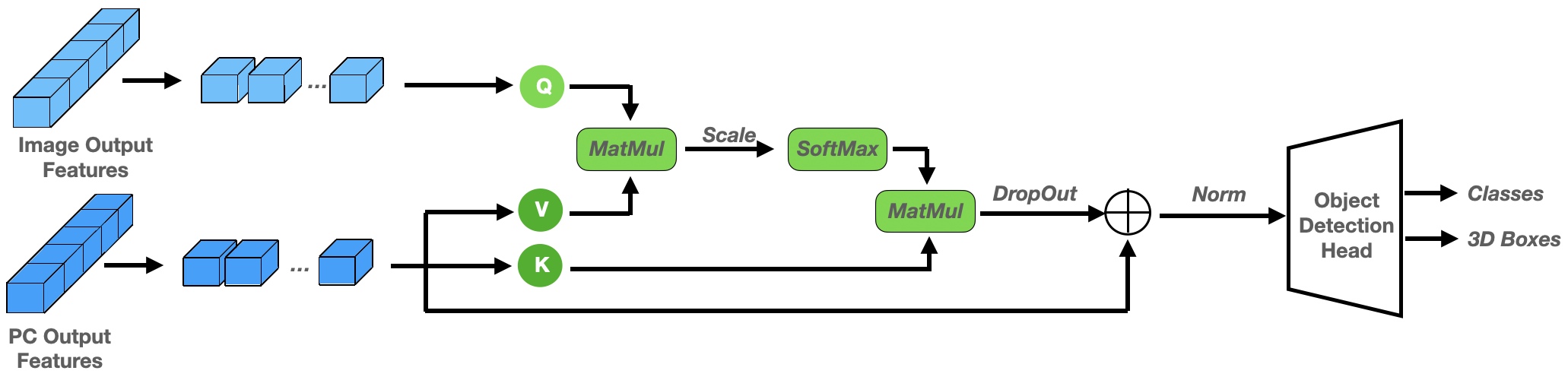}
    \caption{The Cross-Attention Module and Detection Head}
    \label{fig:fig2}
\end{figure*}

\paragraph{Cross-Attention Module.}
Following the decoding of the two feature branches, a method to combine them is needed.
One straightforward way is to simply connect these two feature branches by concatenating them.
This way appears too rough, which may lead the model to suffer from information cuts, resulting in reduced performance (shown in 4.3).
Thus, a multi-head cross-attention mechanism (CA;~\cite{Attention}) is introduced to deeply align and refine these features, which is shown in \cref{fig:fig1}.

Specifically, the output of the point decoder is treated as the source of \(\mathbf{k}\) and \(\mathbf{v}\), whereas the output of the image decoder is projected onto \(\mathbf{q}\).
The cross-attention process of the two-stream features is formulated as follows:
\begin{equation}
    \text{CA}(\mathbf{q}, \mathbf{k}, \mathbf{v}) = \operatorname{Attn}\left(\mathbf{q w_i^q, k w_i^k, v w_i^v}\right) \mathbf{w}^{\text{out}}
\end{equation}
where
\begin{equation}
    \operatorname{Attn}(\mathbf{q}, \mathbf{k}, \mathbf{v}) = \operatorname{Softmax}\left(\frac{\mathbf{q} \mathbf{k}^T}{\sqrt{d_k}}\right) \mathbf{v}
\end{equation}
such that \(\mathbf{q}\), \(\mathbf{k}\), and \(\mathbf{v}\) are linearly projected to compute the attention matrix~\cite{Attention}, and \(\mathbf{q_i^q}\), \(\mathbf{w_i^k}\) and \(\mathbf{w_i^v}\) are the projection layers with shapes \(\mathbb{R}^{d_{\text{model}} \times d_q}\), \(\mathbb{R}^{d_{\text{model}} \times d_k}\) and \(\mathbb{R}^{d_{\text{model}} \times d_v}\).
Then, the refined BEV feature is obtained from the output layer \(\textbf{w}^{\text{Out}} \in \mathbb{R}^{d_v \times d_{\text{model}}}\).
In addition, a residual path from the original point encoder is added to enhance feature propagation and strengthen the model's flexibility:
\begin{equation}
    \mathbf{x} = \text{CA}(\mathbf{q, k, v}) + \mathbf{x}
\end{equation}
Seeking to learn more effectively, these aligned RoI features will be delivered to the detection head and produce results for bounding box regression and classification.

\subsection{Training and Inference}%
\label{sec:training-inference}

\paragraph{Object Detection Head and Loss Function.}
Once getting the Transformer output, we add a bounding box head and a classification head to the output of the CA module. The regression output features can be represented as $\mathbf{\hat{u}^\textit{i}} = (cx, cy, cz, l, w, h, \theta)$, while the classification head outputs $\mathbf{\hat{V}} \in \mathbb{R}^{N \times (C + 1)}$ features, where C is the number of classes. One class is added for the \enquote{no object} class, indicating the absence of any recognized object. Each $\mathbf{\hat{v}^\textit{i}}$ is the predicted probability of the object belonging to the positive class.
We set the ground truth bounding box features to be $\mathbf{U}$ and ground truth class features to be $\mathbf{V}$, where each $\mathbf{v^\textit{i}}$ is a one-hot encoder setting the component corresponding to the class label to be 1 while zeroing all other components.

The Hungarian set prediction loss is applied following~\cite{DETR, SparseRCNN, DiffusionDet}. In addition, by choosing the top \(k\) predictions with the lowest cost using OTA~\cite{OTA}, an optimum transport assignment approach, we assign numerous predictions to each ground truth. For the cost computing on both of the Hungarian set prediction and OTA, we utilize Focal Loss~\cite{FocalLoss} as classification cost, \(\ell_1\) Loss, Generalized IoU Loss~\cite{giou} and Center Loss~\cite{DeformableDETR} as regression cost. Formally:
\begin{equation}
\begin{split}
    \mathcal{L}_{\text{Total}} = {\lambda_1}\mathcal{L}_{\text{cls}} + {\lambda_2}\mathcal{L}_{\text{reg}}.
\end{split}
\end{equation}

The \textbf{classification loss} measures the error in predicting the object class label.
Here, the Focal Loss function~\cite{FocalLoss} is a modified version of the Cross-Entropy:
\begin{equation}
    \mathcal{L}_{\text{cls}} = -\sum_{i} [\mathbf{v^\textit{i}} (1-\hat{\mathbf{v^\textit{i}}})^\gamma \log(\hat{\mathbf{v^\textit{i}}}) + (1-\mathbf{v^\textit{i}}) \hat{\mathbf{v^\textit{i}}}^\gamma \log(1-\hat{\mathbf{v^\textit{i}}})],
\end{equation}
where $\gamma$ is a modulating factor that controls the weight given to each example.
The \textbf{regression loss} measures the error in predicting the object's bounding box location (including center, size, and heading). Inspired by \citet{DeformableDETR}, we add the center loss $\mathcal{L}_{\text{center}}$ to jointly optimize for the best 3D bounding box estimation under the 3D IoU metric:
\begin{equation}
\begin{split}
    \mathcal{L}_{\text{reg}} = {\lambda_3}\mathcal{L}_{\text{L1}} + {\lambda_4}\mathcal{L}_{\text{Giou}} + {\lambda_5}\mathcal{L}_{\text{Center}} 
\end{split}
\end{equation}

\paragraph{Training.}
During training, we build the diffusion process from ground truth to noise filters relying on the corresponding bounding boxes, teaching the model to reverse this procedure after the noise has been added.
To perturb these ground truth boxes to noisy ones with \cref{eq:diffusion-process}, we randomly choose a time step $t$.
Additionally, noisy box features are used to create noisy instance filters for training.
The last step consists of padding the ground truth bounding boxes and their corruption as per \citet{DiffusionDet}.
Thus, we obtain the predicted 3D bounding boxes:
\begin{equation}%
\label{eq:diffusion-process}
    \begin{aligned}
        \mathbf{u}_t & =\sqrt{\bar{\alpha}_t} \mathbf{u}_0+\left(1-\bar{\alpha}_t\right) \epsilon \\
        \mathbf{\theta}_0 & =\eta\left(f\left(\mathbf{u}_t, t\right)\right)
    \end{aligned}
\end{equation}
where $\eta$ denotes the 3D detection head layer and $f(\mathbf{u}, t))$ is the denoising process of the decoder.

\paragraph{Inference.}
Denoising sampling from noise to instance filters makes up the {\ThreeDifFusionDet} inference pipeline.
We first start with bounding boxes $\mathbf{u}_T$ sampled from a Gaussian distribution, where the model iteratively uses to refine its predictions as follows:
\begin{equation}
    \begin{aligned}
        & \mathbf{u}_0=f\left(\cdots\left(f\left(\mathbf{u}_{T-s}, T-s\right)\right)\right) \quad s=\{0, \cdots, T\}, \\
        & \mathbf{\theta}_0=\eta\left(\mathbf{u}_0\right),
    \end{aligned}
\end{equation}
Note that DDIM\cite{DDIM} is also used in our model, in line with DiffusionDet.


\section{Experiments}

\subsection{Experimental Setup}

\paragraph{Datasets.}
To demonstrate the effectiveness of {\ThreeDifFusionDet}, we present results on the KITTI 3D object detection benchmark~\cite{Kitti}.
It is divided into 7,481 training samples and 7,518 testing samples.
The training samples are commonly divided into a training set (3,712 samples) and a validation set (3,769 samples) following~\cite{KITTIdataSplit}, which we adopt.
Before being fed into the models, the dataset is augmented by random 3D flips, random rotation, scale, and translation,
and shuffled the point data. 
We compare {\ThreeDifFusionDet} with existing methods on the test set by training our model on both the training and validation sets.
We evaluate the validation set for ablation by training our model on only the train set.

\paragraph{Baselines.}
We compare our model to several baselines: PV-RCNN~\cite{PVRCNN}, MVX-Net~\cite{MVXNet}, PointRCNN~\cite{PointRCNN}, Part-$A^2$-free~\cite{PartA2Point}, CT3D~\cite{ct3d}, and 3D-SSD~\cite{3DSSD}.
All of them have been popular high-performance 3D object detection frameworks in recent years. 

\paragraph{Implementation Details.}
We implement {\ThreeDifFusionDet} using the MMDetection3D library~\cite{MMDet3D}.
The model is optimized using the Adam~\cite{Adam} optimizer with a learning rate of 0.0001, optimizer momentum $(\beta_1, \beta_2) = (0.9, 0.999)$, and a dropout rate of 0.3.
We train the model on an NVIDIA RTX A6000 GPU for 60 epochs and validate after each epoch.
The voxel grid is defined by a range and voxel size in 3D space.
On KITTI, we use $[2,46.8] \times[-30.08,30.08] \times[-3,1]$ for the range and $[0.16,0.16,0.16]$ for the voxel size for the $x, y$, and $z$ axes, respectively.
For the direct set prediction, we set the number of proposal boxes to 300.
For the diffusion model, we use Gaussian diffusion with \(D = 1000\).
Following \citet{DeformableDETR, DETR, DiffusionDet}, we set the loss weights ${\lambda_3} = 2.5$ while letting others be 1.


\begin{table*}[!htb]
    \centering
    \begin{tabular}{lccccccc}
        \toprule
        \multirow{2}{*}{Model} & \multicolumn{3}{c}{$\text{mAP}_{\text{BEV}}$ ($\text{IoU} = 0.7$)} & \multicolumn{3}{c}{$\text{mAP}_{\text{3D}}$ ($\mathrm{IoU}=0.7$)} \\
        \cmidrule(lr){2-4}\cmidrule(lr){5-7} & Easy & Med & Hard & Easy & Med & Hard \\
        \midrule
        PV-RCNN                       & 86.2 & 84.8 & 78.7 &	N/A	& N/A  & N/A \\
        MVX-Net (VF)                  & 88.6 & 84.6 & 78.6 & 82.3 & 72.2 & 66.8 \\
        MVX-Net (PF)                  & 89.5 & 84.9 & 79.0 & 85.5 & 73.3 & 67.4  \\
        PointRCNN                     & 87.9 & 90.2 & 85.5 & 88.6 & 88.9 & 77.4 \\
        CT3D                          & 88.5 & 86.1 & 79.0 & 90.5 & 87.1 & 79.0 \\
        3D-SSD                        & 89.7 & 89.5 & 78.7 & 89.4 & 87.5 & 78.4 \\
        Part-$A^2$-free                  & 88.0 & 90.2 & \textbf{85.9} & 89.0 & \textbf{88.5} & 78.4 \\
        3DifFusionDet (\(D = 4\)) &\textbf{89.9} & \textbf{91.3} & 85.3          & \textbf{90.5} &   88.2 & \textbf{79.7} \\
        3DifFusionDet (\(D = 8\)) &\textbf{90.3} & \textbf{91.8} & \textbf{86.3} & \textbf{91.3} & \textbf{89.5} & \textbf{80.4} \\
        \bottomrule
    \end{tabular}%
    \caption{Comparison of attained validation mAP (in \%) on KITTI with $\text{IoU} = 0.7$.}
    \label{tab:results-kitti}
\end{table*}


\begin{table}[!htb]
    \centering
    \begin{tabular}{lccc}
        \toprule
        & Easy & Med & Hard \\
        \midrule
        W/o Image RoI Align & 86.4 & 75.0 & 74.8 \\
        W/o encoder feature fusion & 87.0 & 77.8 & 75.4 \\
        W/ Both & \textbf{90.5} & \textbf{88.2} & \textbf{79.7} \\
        \bottomrule
    \end{tabular}
    \caption{Performance gap on whether using Image RoI Align on $\text{AP}_{\text{3D}}(\text{IoU} = 0.7)$}
    \label{tab:perf-gap}
\end{table}

\begin{table}[htb]
    \centering
    \begin{tabular}{lccc}
        \toprule
        Fusion Align & Easy & Med & Hard \\
        \midrule
        Sum & 86.8 & 72.0 & 70.8 \\
        Concat & 86.5 & 72.4 & 71.1 \\
        DP & 86.2 & 80.5 & 75.6 \\
        MLP & 87.0 & 83.9 & 76.5 \\
        CA & 88.8 & 87.9 & 78.1 \\
        Res-CA & \textbf{90.5} & \textbf{88.2} & \textbf{79.7} \\
        \bottomrule
    \end{tabular}
    \caption{Feature align strategies on $\text{AP}_{\text{3D}}$ ($\text{IoU} = 0.7$).}
    \label{tab:align-strategies}
\end{table}

\begin{table}[htb]
    \centering
    \begin{tabular}{cccccc}
        \toprule
        \#Boxes & \(D\) & Easy & Med & Hard & FPS \\
        \midrule 
        300 & 1 & 87.1 & 85.3 & 77.7 & \textbf{1.9} \\
        100 & 4 & 86.8 & 85.5 & 77.3 & 1.3 \\
        300 & 4 & 90.5 & 88.2 & 79.7 & 1.2 \\  
        300 & 8 & \textbf{91.3} & \textbf{89.5} & \textbf{80.4} & 0.6 \\  
        \bottomrule
    \end{tabular}
    \caption{Accuracy vs.\ speed.
    Using more proposal boxes incurs a higher performance gain at the cost of latency on $\text{AP}_{\text{3D}}$ ($\text{IoU} = 0.7$).}
    \label{tab:accuracy-vs-speed}
\end{table}

\subsection{Results}

We conduct experiments on the KITTI 3D object detection benchmark.
Following the standard KITTI evaluation protocol ($\text{IoU} = 0.7$) for measuring the detection performance, \cref{tab:results-kitti} shows the mean average precision (mAP) scores for the {\ThreeDifFusionDet} method compared to the state-of-the-art methods on the KITTI validation set using 3D and bird’s eye view (BEV) evaluation. 
We report its performance for \(D \in \{4, 8\}\), following[diffusionDet, diffusionist] and bold-face the two best-performing models for each task.

As per \cref{tab:results-kitti}, our approach shows significant performance improvements compared to the baselines.
With \(D = 4\), it is able to outperform most of the baselines with a relatively short inference time.
By further increasing \(D\) such that \(D = 8\), taking into account a higher inference time, we achieve the best performance among all models.
This flexibility reveals a wide range of potential usages. The trade-off between accuracy and inference speed is discussed in \cref{sec:ablation-studies}.

\subsection{Ablation Studies}%
\label{sec:ablation-studies}

Firstly, we show the necessity of keeping the \textbf{Image RoI Align branch} and the \textbf{encoder feature fusion}. 
For designing a 3D Object detector from both Camera and Lidar using the diffusion model, the most straightforward approach should be directly applying the generated noisy 3D boxes as input to the fused 3D features.
This way, however, could suffer from information cut, which leads to reduced performance, as shown in \cref{tab:perf-gap}.
Drawing upon it, besides putting the point cloud RoIAlign under the encoded 3D features, we also create the second branch that makes the image RoIAlign under the encoded 2D features. Much better utilization of the complementing information offered by the two modalities is shown by significantly higher performance.

Then we analyze the influence of using different \textbf{fusion strategies}: given the learned 2D and 3D represented features, how to combine them more efficiently. 

Compared to 2D features, 3D features contain an additional dimension, making them more challenging to learn. Inspired by \citet{MVXNet}, we add an information flow path from image feature to point feature by additionally projecting points from the LiDAR sensors, using the concatenation of image features and the corresponding points to be jointly processed by the VoxelNet architecture.
\Cref{tab:align-strategies} shows its benefits gain on detection accuracy. 

Another part that needs fusion is the connection of the two feature branches following the decoding. Here we applied a multi-head cross-attention mechanism~\cite{Attention} to deeply align and refine these features. Besides this way, more straightforward ways like using concatenation operation, sum operation, direct product operation (DP;~\cite{DirectProduct}), and using a multi-layer perceptron (MLP) are all investigated.
\Cref{tab:accuracy-vs-speed} shows the results. Among all, the cross-attention mechanism shows the best performance, with almost the same training and inference speed.
This demonstrates its promising features.

Next, we investigate the \textbf{accuracy and inference speed trade-off}.

By comparing the 3D detection accuracy and number of frames per second (FPS), we show the influence of choosing different proposal boxes as well as \(D\).
The number of proposal boxes are chosen from \({100, 300}\), whereas \(D\) is chosen from \({1, 4, 8}\).
The run time is evaluated on a single NVIDIA RTX A6000 GPU with a batch size of 1.
We see that increasing the number of proposal boxes from 100 to 300 significantly increases the accuracy gain with negligible latency cost (1.3 FPS vs.\ 1.2 FPS).
On the other hand, better detection accuracy incurs a longer inference time.
As we vary the \(D\) from 1 to 8, the 3D detection accuracy increases from steeply (Easy: 87.1 mAP to 90.5 mAP) to relatively slowly (Easy: 90.5 AP to 91.3 mAP), while the FPS decreases constantly. Note that we did the reference on only one RTX A6000 GPU.
Referred to \cite{Diffusioninst}, who got nearby FPS values as ours by running on a single V100 GPU, arguing that their method could reach a similar real-time level as \cite{DiffusionDet} in 2D object detections. Our model should also reach real-time inference speed when running on a high-performance GPU like A100, which \cite{DiffusionDet} utilized.

\subsection{Case Study and Future Work}
\label{sec:future-work}

Based on its unique properties, we discuss potential usages of our {\ThreeDifFusionDet}. Generally, inferring accurately, robustly, and in real-time are three requirements in object detection tasks~\cite{3DODinAD}. In the perception field of self-driving cars, considering the fact that a high-speed car needs to take extra time and distance to slow down or change direction due to inertia, the perception model is especially sensitive to the real-time requirement. More importantly, to guarantee a comfortable riding experience, the car should run as smoothly as possible with the minimum absolute value of acceleration, under the premise of safety. It is one primary advantage over other similar self-driving car products that withing smoother riding experiences. To this end, whether it is to speed up, slow down, or turn, self-driving cars should start to respond in a quick manner. The quicker a car starts to respond, the more robust space it earns for following operations and adjustments. This is even more important compared to getting the most precise detected object's classification or location first: as a car starts to respond, there is still time and distance to adjust its manners, which could be utilized to make further inferences in a more precise way, whose result subsequently finetunes the driving operation of the car. 

Our {\ThreeDifFusionDet} naturally matches the need. As \cref{tab:accuracy-vs-speed} shows, when the inference step is small, the model could make inferences in a quick time with roughly high-accurate results. This initial perception is precise enough for a self-driving car to start its new response. As the inference step grows, higher accurate detected objects are generated, further finetuning its response. This \textit{progressive detection manner} fits in nicely for this task. In addition, since our model could change the number of proposal boxes during reference, the prior information obtained from the small steps could in turn be used to optimize the real-time proposal box number. As \cref{tab:accuracy-vs-speed} shows, the performance varies under different prior proposal boxes. Therefore, developing such a self-adaptive detector is one bright future work. 

Besides self-driving cars, our model inherently matches any real-world scenario that requires short inferring time in continuous reaction space, especially in scenes where the detector is moving based on the detected result. Benefits from the nature of the diffusion model, {\ThreeDifFusionDet} could find the almost-accurate real space regions of interest in a quick time, triggering the machine to start making new operations and self-optimization. The following higher-accurate perceptron further finetunes the machine's operations. To deploy our model into these moving detectors, one open issue is the strategy that combines the inferred information from between larger steps' early infers and smaller steps' latest infer, which is another open question.






\section{Conclusion}

This paper presents 3DifFusionDet, a novel 3D object detector with robust LiDAR and camera fusion.
Formulating 3D object detection as a generative denoising process, it is the first work that applies a diffusion model to 3D object detection.
This work investigates the most effective camera-LiDAR fusion alignment strategies within the context of the generative denoising process framework, and it proposes fusion align strategies to make full use of the complementing information offered by the two modalities.
3DifFusionDet achieves favorable performance compared to well-established detectors, demonstrating a promising future of diffusion models in object detection tasks.
The robust learning results and flexible inference mode makes it promising in potential usages.


\bibliography{aaai24}

\begin{thebibliography}{74}
\providecommand{\natexlab}[1]{#1}

\bibitem[{Amit et~al.(2022)Amit, Shaharbany, Nachmani, and Wolf}]{SegDiff}
Amit, T.; Shaharbany, T.; Nachmani, E.; and Wolf, L. 2022.
\newblock SegDiff: Image Segmentation with Diffusion Probabilistic Models.
\newblock arXiv:2112.00390.

\bibitem[{Asiedu et~al.(2022)Asiedu, Kornblith, Chen, Parmar, Minderer, and Norouzi}]{DecoderDenoisingPretrainingForSemanticSegmentation}
Asiedu, E.~B.; Kornblith, S.; Chen, T.; Parmar, N.; Minderer, M.; and Norouzi, M. 2022.
\newblock Decoder Denoising Pretraining for Semantic Segmentation.
\newblock arXiv:2205.11423.

\bibitem[{Austin et~al.(2023{\natexlab{a}})Austin, Johnson, Ho, Tarlow, and van~den Berg}]{StructuredDenoisingDiffusionModel}
Austin, J.; Johnson, D.~D.; Ho, J.; Tarlow, D.; and van~den Berg, R. 2023{\natexlab{a}}.
\newblock Structured Denoising Diffusion Models in Discrete State-Spaces.
\newblock arXiv:2107.03006.

\bibitem[{Austin et~al.(2023{\natexlab{b}})Austin, Johnson, Ho, Tarlow, and van~den Berg}]{StructuredDenoisingDiffusionModelsinDiscreteState_Spaces}
Austin, J.; Johnson, D.~D.; Ho, J.; Tarlow, D.; and van~den Berg, R. 2023{\natexlab{b}}.
\newblock Structured Denoising Diffusion Models in Discrete State-Spaces.
\newblock arXiv:2107.03006.

\bibitem[{Avrahami, Lischinski, and Fried(2022)}]{BlendedDiffusionforText_drivenEditing}
Avrahami, O.; Lischinski, D.; and Fried, O. 2022.
\newblock Blended Diffusion for Text-driven Editing of Natural Images.
\newblock In \emph{2022 {IEEE}/{CVF} Conference on Computer Vision and Pattern Recognition ({CVPR})}. {IEEE}.

\bibitem[{Bai et~al.(2022)Bai, Hu, Zhu, Huang, Chen, Fu, and Tai}]{TransFusion}
Bai, X.; Hu, Z.; Zhu, X.; Huang, Q.; Chen, Y.; Fu, H.; and Tai, C.-L. 2022.
\newblock TransFusion: Robust LiDAR-Camera Fusion for 3D Object Detection With Transformers.
\newblock In \emph{Proceedings of the IEEE/CVF Conference on Computer Vision and Pattern Recognition (CVPR)}, 1090--1099.

\bibitem[{Baranchuk et~al.(2022)Baranchuk, Rubachev, Voynov, Khrulkov, and Babenko}]{EfficientSS}
Baranchuk, D.; Rubachev, I.; Voynov, A.; Khrulkov, V.; and Babenko, A. 2022.
\newblock Label-Efficient Semantic Segmentation with Diffusion Models.
\newblock arXiv:2112.03126.

\bibitem[{Cao et~al.(2016)Cao, Zhou, Li, and Yu}]{DirectProduct}
Cao, B.; Zhou, H.; Li, G.; and Yu, P.~S. 2016.
\newblock Multi-view Machines.
\newblock In \emph{Proceedings of the Ninth {ACM} International Conference on Web Search and Data Mining}. {ACM}.

\bibitem[{Carion et~al.(2020)Carion, Massa, Synnaeve, Usunier, Kirillov, and Zagoruyko}]{DETR}
Carion, N.; Massa, F.; Synnaeve, G.; Usunier, N.; Kirillov, A.; and Zagoruyko, S. 2020.
\newblock End-to-End Object Detection with Transformers.

\bibitem[{Chen et~al.(2022{\natexlab{a}})Chen, Sun, Song, and Luo}]{DiffusionDet}
Chen, S.; Sun, P.; Song, Y.; and Luo, P. 2022{\natexlab{a}}.
\newblock {DiffusionDet}: Diffusion Model for Object Detection.
\newblock \emph{arXiv preprint arXiv:2211.09788}.

\bibitem[{Chen et~al.(2022{\natexlab{b}})Chen, Li, Saxena, Hinton, and Fleet}]{AGeneralistFramework}
Chen, T.; Li, L.; Saxena, S.; Hinton, G.; and Fleet, D.~J. 2022{\natexlab{b}}.
\newblock A Generalist Framework for Panoptic Segmentation of Images and Videos.
\newblock arXiv:2210.06366.

\bibitem[{Chen et~al.(2017)Chen, Kundu, Zhu, Ma, Fidler, and Urtasun}]{KITTIdataSplit}
Chen, X.; Kundu, K.; Zhu, Y.; Ma, H.; Fidler, S.; and Urtasun, R. 2017.
\newblock 3D Object Proposals using Stereo Imagery for Accurate Object Class Detection.
\newblock arXiv:1608.07711.

\bibitem[{Chen et~al.(2016)Chen, Ma, Wan, Li, and Xia}]{MV3D}
Chen, X.; Ma, H.; Wan, J.; Li, B.; and Xia, T. 2016.
\newblock Multi-View 3D Object Detection Network for Autonomous Driving.

\bibitem[{Chen et~al.(2023)Chen, Liu, Zhang, Qi, and Jia}]{VoxelNeXt}
Chen, Y.; Liu, J.; Zhang, X.; Qi, X.; and Jia, J. 2023.
\newblock VoxelNeXt: Fully Sparse VoxelNet for 3D Object Detection and Tracking.
\newblock arXiv:2303.11301.

\bibitem[{Chitta et~al.(2022)Chitta, Prakash, Jaeger, Yu, Renz, and Geiger}]{TransFuser}
Chitta, K.; Prakash, A.; Jaeger, B.; Yu, Z.; Renz, K.; and Geiger, A. 2022.
\newblock TransFuser: Imitation with Transformer-Based Sensor Fusion for Autonomous Driving.

\bibitem[{Dhariwal and Nichol(2021)}]{DiffusionModelsBeatGANs}
Dhariwal, P.; and Nichol, A. 2021.
\newblock Diffusion Models Beat GANs on Image Synthesis.
\newblock arXiv:2105.05233.

\bibitem[{Fan et~al.(2021)Fan, Pang, Zhang, Wang, Zhao, Wang, Wang, and Zhang}]{SST}
Fan, L.; Pang, Z.; Zhang, T.; Wang, Y.-X.; Zhao, H.; Wang, F.; Wang, N.; and Zhang, Z. 2021.
\newblock Embracing Single Stride 3D Object Detector with Sparse Transformer.
\newblock arXiv:2112.06375.

\bibitem[{Fan et~al.(2022)Fan, Wang, Wang, and Zhang}]{FSD++}
Fan, L.; Wang, F.; Wang, N.; and Zhang, Z. 2022.
\newblock Fully Sparse 3D Object Detection.
\newblock arXiv:2207.10035.

\bibitem[{Gao et~al.(2021)Gao, Zheng, Wang, Dai, and Li}]{SMCA}
Gao, P.; Zheng, M.; Wang, X.; Dai, J.; and Li, H. 2021.
\newblock Fast Convergence of DETR with Spatially Modulated Co-Attention.
\newblock arXiv:2101.07448.

\bibitem[{Ge et~al.(2021)Ge, Liu, Li, Yoshie, and Sun}]{OTA}
Ge, Z.; Liu, S.; Li, Z.; Yoshie, O.; and Sun, J. 2021.
\newblock OTA: Optimal Transport Assignment for Object Detection.
\newblock arXiv:2103.14259.

\bibitem[{Geiger, Lenz, and Urtasun(2012)}]{Kitti}
Geiger, A.; Lenz, P.; and Urtasun, R. 2012.
\newblock Are we ready for Autonomous Driving? The {KITTI} Vision Benchmark Suite.
\newblock In \emph{2012 IEEE Conference on Computer Vision and Pattern Recognition}, 3354--3361. IEEE.

\bibitem[{Girshick(2015)}]{Fast_RCNN}
Girshick, R. 2015.
\newblock Fast R-CNN.

\bibitem[{Girshick et~al.(2013)Girshick, Donahue, Darrell, and Malik}]{RichFeatureHierarchies}
Girshick, R.~B.; Donahue, J.; Darrell, T.; and Malik, J. 2013.
\newblock Rich feature hierarchies for accurate object detection and semantic segmentation.
\newblock \emph{CoRR}, abs/1311.2524.

\bibitem[{Gomez-Ojeda, Briales, and Gonzalez-Jimenez(2016)}]{PL-SVO}
Gomez-Ojeda, R.; Briales, J.; and Gonzalez-Jimenez, J. 2016.
\newblock PL-SVO: Semi-direct Monocular Visual Odometry by combining points and line segments.
\newblock In \emph{2016 IEEE/RSJ International Conference on Intelligent Robots and Systems (IROS)}, 4211--4216.

\bibitem[{Graikos et~al.(2023)Graikos, Malkin, Jojic, and Samaras}]{graikos2023diffusion}
Graikos, A.; Malkin, N.; Jojic, N.; and Samaras, D. 2023.
\newblock Diffusion models as plug-and-play priors.
\newblock arXiv:2206.09012.

\bibitem[{Gu et~al.(2022)Gu, Chen, Xu, Lan, Meng, and Wang}]{Diffusioninst}
Gu, Z.; Chen, H.; Xu, Z.; Lan, J.; Meng, C.; and Wang, W. 2022.
\newblock DiffusionInst: Diffusion Model for Instance Segmentation.
\newblock arXiv:2212.02773.

\bibitem[{Gupta and Lam(1998)}]{l2Loss}
Gupta, A.; and Lam, S.~M. 1998.
\newblock Weight decay backpropagation for noisy data.
\newblock \emph{Neural networks}, 11(6): 1127--1138.

\bibitem[{He et~al.(2017)He, Gkioxari, Dollár, and Girshick}]{Mask_RCNN}
He, K.; Gkioxari, G.; Dollár, P.; and Girshick, R. 2017.
\newblock Mask R-CNN.

\bibitem[{He et~al.(2015)He, Zhang, Ren, and Sun}]{ResNet}
He, K.; Zhang, X.; Ren, S.; and Sun, J. 2015.
\newblock Deep Residual Learning for Image Recognition.
\newblock arXiv:1512.03385.

\bibitem[{Ho, Jain, and Abbeel(2020{\natexlab{a}})}]{DDPM}
Ho, J.; Jain, A.; and Abbeel, P. 2020{\natexlab{a}}.
\newblock Denoising Diffusion Probabilistic Models.
\newblock arXiv:2006.11239.

\bibitem[{Ho, Jain, and Abbeel(2020{\natexlab{b}})}]{DiffusionModel}
Ho, J.; Jain, A.; and Abbeel, P. 2020{\natexlab{b}}.
\newblock Denoising Diffusion Probabilistic Models.
\newblock \emph{Advances in Neural Information Processing Systems}, 33: 6840--6851.

\bibitem[{Hoogeboom et~al.(2022)Hoogeboom, Satorras, Vignac, and Welling}]{DiffusionforMoleculeGeneration}
Hoogeboom, E.; Satorras, V.~G.; Vignac, C.; and Welling, M. 2022.
\newblock Equivariant Diffusion for Molecule Generation in 3D.
\newblock arXiv:2203.17003.

\bibitem[{Jin et~al.(2023)Jin, Li, Cheng, Li, Ji, Liu, Yuan, and Chen}]{DiffusionRet}
Jin, P.; Li, H.; Cheng, Z.; Li, K.; Ji, X.; Liu, C.; Yuan, L.; and Chen, J. 2023.
\newblock DiffusionRet: Generative Text-Video Retrieval with Diffusion Model.
\newblock arXiv:2303.09867.

\bibitem[{Kingma and Ba(2014)}]{Adam}
Kingma, D.~P.; and Ba, J. 2014.
\newblock {Adam}: A Method for Stochastic Optimization.
\newblock \emph{arXiv preprint arXiv:1412.6980}.

\bibitem[{Kong et~al.(2021)Kong, Ping, Huang, Zhao, and Catanzaro}]{DiffWave}
Kong, Z.; Ping, W.; Huang, J.; Zhao, K.; and Catanzaro, B. 2021.
\newblock DiffWave: A Versatile Diffusion Model for Audio Synthesis.
\newblock arXiv:2009.09761.

\bibitem[{Ku et~al.(2017)Ku, Mozifian, Lee, Harakeh, and Waslander}]{AVOD}
Ku, J.; Mozifian, M.; Lee, J.; Harakeh, A.; and Waslander, S. 2017.
\newblock Joint 3D Proposal Generation and Object Detection from View Aggregation.

\bibitem[{Li et~al.(2022{\natexlab{a}})Li, Zhang, Liu, Guo, Ni, and Zhang}]{DN-DETR}
Li, F.; Zhang, H.; Liu, S.; Guo, J.; Ni, L.~M.; and Zhang, L. 2022{\natexlab{a}}.
\newblock DN-DETR: Accelerate DETR Training by Introducing Query DeNoising.
\newblock arXiv:2203.01305.

\bibitem[{Li et~al.(2022{\natexlab{b}})Li, Thickstun, Gulrajani, Liang, and Hashimoto}]{Diffusion_LMImprovesControllableTextGeneration}
Li, X.~L.; Thickstun, J.; Gulrajani, I.; Liang, P.; and Hashimoto, T.~B. 2022{\natexlab{b}}.
\newblock Diffusion-LM Improves Controllable Text Generation.
\newblock arXiv:2205.14217.

\bibitem[{Lin et~al.(2017{\natexlab{a}})Lin, Dollár, Girshick, He, Hariharan, and Belongie}]{FPN}
Lin, T.-Y.; Dollár, P.; Girshick, R.; He, K.; Hariharan, B.; and Belongie, S. 2017{\natexlab{a}}.
\newblock Feature Pyramid Networks for Object Detection.
\newblock arXiv:1612.03144.

\bibitem[{Lin et~al.(2017{\natexlab{b}})Lin, Goyal, Girshick, He, and Dollár}]{FocalLoss}
Lin, T.-Y.; Goyal, P.; Girshick, R.; He, K.; and Dollár, P. 2017{\natexlab{b}}.
\newblock Focal Loss for Dense Object Detection.

\bibitem[{Liu et~al.(2022)Liu, Tang, Amini, Yang, Mao, Rus, and Han}]{BEVFusion}
Liu, Z.; Tang, H.; Amini, A.; Yang, X.; Mao, H.; Rus, D.; and Han, S. 2022.
\newblock BEVFusion: Multi-Task Multi-Sensor Fusion with Unified Bird's-Eye View Representation.
\newblock arXiv:2205.13542.

\bibitem[{Mao et~al.(2023)Mao, Shi, Wang, and Li}]{3DODinAD}
Mao, J.; Shi, S.; Wang, X.; and Li, H. 2023.
\newblock 3D Object Detection for Autonomous Driving: A Comprehensive Survey.
\newblock arXiv:2206.09474.

\bibitem[{Misra, Girdhar, and Joulin(2021)}]{3DETR}
Misra, I.; Girdhar, R.; and Joulin, A. 2021.
\newblock An End-to-End Transformer Model for 3D Object Detection.
\newblock In \emph{Proceedings of the IEEE/CVF International Conference on Computer Vision (ICCV)}, 2906--2917.

\bibitem[{MMDetection3D(2020)}]{MMDet3D}
MMDetection3D, C. 2020.
\newblock {MMDetection3D: OpenMMLab} next-generation platform for general {3D} object detection.
\newblock \url{https://github.com/open-mmlab/mmdetection3d}.

\bibitem[{Nie et~al.(2022)Nie, Guo, Huang, Xiao, Vahdat, and Anandkumar}]{DiffusionModelsforAdversarialPurification}
Nie, W.; Guo, B.; Huang, Y.; Xiao, C.; Vahdat, A.; and Anandkumar, A. 2022.
\newblock Diffusion Models for Adversarial Purification.
\newblock arXiv:2205.07460.

\bibitem[{Park, Lee, and Kwon(2022)}]{NeuralMarkovControlled_SDE}
Park, S.~W.; Lee, K.; and Kwon, J. 2022.
\newblock Neural Markov Controlled {SDE}: Stochastic Optimization for Continuous-Time Data.
\newblock In \emph{International Conference on Learning Representations}.

\bibitem[{Park, Lepetit, and Woo(2008)}]{Multiple3DObjectTrackingforAugmentedReality}
Park, Y.; Lepetit, V.; and Woo, W. 2008.
\newblock Multiple 3D Object Tracking for Augmented Reality.
\newblock In \emph{Proceedings of the 7th IEEE/ACM International Symposium on Mixed and Augmented Reality}, ISMAR '08, 117–120. USA: IEEE Computer Society.
\newblock ISBN 9781424428403.

\bibitem[{Prakash, Chitta, and Geiger(2021)}]{MultiModalFusionTransformer}
Prakash, A.; Chitta, K.; and Geiger, A. 2021.
\newblock Multi-Modal Fusion Transformer for End-to-End Autonomous Driving.
\newblock arXiv:2104.09224.

\bibitem[{Qi et~al.(2017)Qi, Liu, Wu, Su, and Guibas}]{FrustumNet}
Qi, C.~R.; Liu, W.; Wu, C.; Su, H.; and Guibas, L.~J. 2017.
\newblock Frustum PointNets for 3D Object Detection from RGB-D Data.

\bibitem[{Qi et~al.(2016)Qi, Su, Mo, and Guibas}]{PointNet}
Qi, C.~R.; Su, H.; Mo, K.; and Guibas, L.~J. 2016.
\newblock PointNet: Deep Learning on Point Sets for 3D Classification and Segmentation.

\bibitem[{Ren et~al.(2015)Ren, He, Girshick, and Sun}]{Faster_RCNN}
Ren, S.; He, K.; Girshick, R.; and Sun, J. 2015.
\newblock Faster R-CNN: Towards Real-Time Object Detection with Region Proposal Networks.

\bibitem[{Rezatofighi et~al.(2019)Rezatofighi, Tsoi, Gwak, Sadeghian, Reid, and Savarese}]{giou}
Rezatofighi, H.; Tsoi, N.; Gwak, J.; Sadeghian, A.; Reid, I.; and Savarese, S. 2019.
\newblock Generalized Intersection over Union: A Metric and A Loss for Bounding Box Regression.
\newblock arXiv:1902.09630.

\bibitem[{Saharia et~al.(2021)Saharia, Ho, Chan, Salimans, Fleet, and Norouzi}]{ImageSuperResolutionviaIterativeRefinement}
Saharia, C.; Ho, J.; Chan, W.; Salimans, T.; Fleet, D.~J.; and Norouzi, M. 2021.
\newblock Image Super-Resolution via Iterative Refinement.
\newblock arXiv:2104.07636.

\bibitem[{Shan et~al.(2023)Shan, Liu, Zhang, Wang, Han, Wang, Ma, and Gao}]{Diffusion_Based3DHumanPoseEstimation}
Shan, W.; Liu, Z.; Zhang, X.; Wang, Z.; Han, K.; Wang, S.; Ma, S.; and Gao, W. 2023.
\newblock Diffusion-Based 3D Human Pose Estimation with Multi-Hypothesis Aggregation.
\newblock arXiv:2303.11579.

\bibitem[{Sheng et~al.(2021)Sheng, Cai, Liu, Deng, Huang, Hua, and Zhao}]{ct3d}
Sheng, H.; Cai, S.; Liu, Y.; Deng, B.; Huang, J.; Hua, X.-S.; and Zhao, M.-J. 2021.
\newblock Improving 3D Object Detection with Channel-wise Transformer.
\newblock arXiv:2108.10723.

\bibitem[{Shi et~al.(2020{\natexlab{a}})Shi, Guo, Jiang, Wang, Shi, Wang, and Li}]{PVRCNN}
Shi, S.; Guo, C.; Jiang, L.; Wang, Z.; Shi, J.; Wang, X.; and Li, H. 2020{\natexlab{a}}.
\newblock {PV-RCNN}: Point-Voxel Feature Set Abstraction for 3D Object Detection.
\newblock In \emph{Proceedings of the IEEE/CVF Conference on Computer Vision and Pattern Recognition}, 10529--10538.

\bibitem[{Shi, Wang, and Li(2018)}]{PointRCNN}
Shi, S.; Wang, X.; and Li, H. 2018.
\newblock PointRCNN: 3D Object Proposal Generation and Detection from Point Cloud.

\bibitem[{Shi et~al.(2020{\natexlab{b}})Shi, Wang, Shi, Wang, and Li}]{PartA2Point}
Shi, S.; Wang, Z.; Shi, J.; Wang, X.; and Li, H. 2020{\natexlab{b}}.
\newblock From Points to Parts: 3D Object Detection from Point Cloud with Part-aware and Part-aggregation Network.
\newblock arXiv:1907.03670.

\bibitem[{Sindagi, Zhou, and Tuzel(2019)}]{MVXNet}
Sindagi, V.~A.; Zhou, Y.; and Tuzel, O. 2019.
\newblock {MVX-Net}: Multimodal VoxelNet for 3D Object Detection.

\bibitem[{Song, Meng, and Ermon(2022)}]{DDIM}
Song, J.; Meng, C.; and Ermon, S. 2022.
\newblock Denoising Diffusion Implicit Models.
\newblock arXiv:2010.02502.

\bibitem[{Song and Ermon(2020)}]{GenerativeModelingByEstimatingGradientsOfTheDataDistribution}
Song, Y.; and Ermon, S. 2020.
\newblock Generative Modeling by Estimating Gradients of the Data Distribution.
\newblock arXiv:1907.05600.

\bibitem[{Sun et~al.(2022)Sun, Tan, Wang, Liu, Xia, Leng, and Anguelov}]{SWFormer}
Sun, P.; Tan, M.; Wang, W.; Liu, C.; Xia, F.; Leng, Z.; and Anguelov, D. 2022.
\newblock SWFormer: Sparse Window Transformer for 3D Object Detection in Point Clouds.
\newblock arXiv:2210.07372.

\bibitem[{Sun et~al.(2020)Sun, Zhang, Jiang, Kong, Xu, Zhan, Tomizuka, Li, Yuan, Wang, and Luo}]{SparseRCNN}
Sun, P.; Zhang, R.; Jiang, Y.; Kong, T.; Xu, C.; Zhan, W.; Tomizuka, M.; Li, L.; Yuan, Z.; Wang, C.; and Luo, P. 2020.
\newblock Sparse {R-CNN:} End-to-End Object Detection with Learnable Proposals.
\newblock \emph{CoRR}, abs/2011.12450.

\bibitem[{Trippe et~al.(2023)Trippe, Yim, Tischer, Baker, Broderick, Barzilay, and Jaakkola}]{DiffusionProbabilisticModelingOfProteinBackbones}
Trippe, B.~L.; Yim, J.; Tischer, D.; Baker, D.; Broderick, T.; Barzilay, R.; and Jaakkola, T. 2023.
\newblock Diffusion probabilistic modeling of protein backbones in 3D for the motif-scaffolding problem.
\newblock arXiv:2206.04119.

\bibitem[{Vaswani et~al.(2017{\natexlab{a}})Vaswani, Shazeer, Parmar, Uszkoreit, Jones, Gomez, Kaiser, and Polosukhin}]{Transformers}
Vaswani, A.; Shazeer, N.; Parmar, N.; Uszkoreit, J.; Jones, L.; Gomez, A.~N.; Kaiser, L.; and Polosukhin, I. 2017{\natexlab{a}}.
\newblock Attention Is All You Need.

\bibitem[{Vaswani et~al.(2017{\natexlab{b}})Vaswani, Shazeer, Parmar, Uszkoreit, Jones, Gomez, Kaiser, and Polosukhin}]{Attention}
Vaswani, A.; Shazeer, N.; Parmar, N.; Uszkoreit, J.; Jones, L.; Gomez, A.~N.; Kaiser, {\L}.; and Polosukhin, I. 2017{\natexlab{b}}.
\newblock Attention is All You Need.
\newblock \emph{Advances in Neural Information Processing Systems}, 30.

\bibitem[{Wang et~al.(2022)Wang, Lyu, Lin, Dai, and Fu}]{GuidedDiffusionModelforAdversarialPurification}
Wang, J.; Lyu, Z.; Lin, D.; Dai, B.; and Fu, H. 2022.
\newblock Guided Diffusion Model for Adversarial Purification.
\newblock arXiv:2205.14969.

\bibitem[{Xiang et~al.(2016)Xiang, Kim, Chen, Ji, Choy, Su, Mottaghi, Guibas, and Savarese}]{ObjectNet3D}
Xiang, Y.; Kim, W.; Chen, W.; Ji, J.; Choy, C.~B.; Su, H.; Mottaghi, R.; Guibas, L.~J.; and Savarese, S. 2016.
\newblock ObjectNet3D: A Large Scale Database for 3D Object Recognition.
\newblock In \emph{European Conference on Computer Vision}.

\bibitem[{Yan, Mao, and Li(2018)}]{SECOND}
Yan, Y.; Mao, Y.; and Li, B. 2018.
\newblock SECOND: Sparsely Embedded Convolutional Detection.
\newblock \emph{Sensors (Basel, Switzerland)}, 18.

\bibitem[{Yang et~al.(2020)Yang, Sun, Liu, and Jia}]{3DSSD}
Yang, Z.; Sun, Y.; Liu, S.; and Jia, J. 2020.
\newblock {3DSSD}: Point-based 3D Single Stage Object Detector.
\newblock In \emph{Proceedings of the IEEE/CVF Conference on Computer Vision and Pattern Recognition}, 11040--11048.

\bibitem[{Zeng et~al.(2022)Zeng, Zhang, Wang, Miao, Liu, Zhan, Hao, and Ma}]{LIFT}
Zeng, Y.; Zhang, D.; Wang, C.; Miao, Z.; Liu, T.; Zhan, X.; Hao, D.; and Ma, C. 2022.
\newblock LIFT: Learning 4D LiDAR Image Fusion Transformer for 3D Object Detection.
\newblock In \emph{2022 IEEE/CVF Conference on Computer Vision and Pattern Recognition (CVPR)}, 17151--17160.

\bibitem[{Zhou et~al.(2023)Zhou, Liu, Hu, Zhou, and Ma}]{UniDistill}
Zhou, S.; Liu, W.; Hu, C.; Zhou, S.; and Ma, C. 2023.
\newblock UniDistill: A Universal Cross-Modality Knowledge Distillation Framework for 3D Object Detection in Bird's-Eye View.
\newblock arXiv:2303.15083.

\bibitem[{Zhou and Tuzel(2017)}]{VoxelNet}
Zhou, Y.; and Tuzel, O. 2017.
\newblock VoxelNet: End-to-End Learning for Point Cloud Based 3D Object Detection.

\bibitem[{Zhu et~al.(2020)Zhu, Su, Lu, Li, Wang, and Dai}]{DeformableDETR}
Zhu, X.; Su, W.; Lu, L.; Li, B.; Wang, X.; and Dai, J. 2020.
\newblock Deformable DETR: Deformable Transformers for End-to-End Object Detection.

\end{thebibliography}

\end{document}